\documentclass[preprint, review, compress,12pt]{elsarticle}

\usepackage{setspace}
\usepackage{graphicx,fancyhdr,amssymb}
\usepackage{color}
\usepackage{times,amsmath}
\usepackage[ps]{skak}
\usepackage{latexsym}
\usepackage{url,doi}
\usepackage{appendix}

\usepackage[linesnumbered,ruled]{algorithm2e}

\usepackage{lineno}
\usepackage{tabularx} 

\usepackage{pdflscape}
\usepackage{fancyhdr}

\usepackage{longtable}
\usepackage{rotating}
\usepackage[left=1in, right=1in, top=1in, bottom=1in]{geometry}
\usepackage{booktabs} 
\usepackage{cleveref} 
\usepackage{tablefootnote}
\usepackage{makecell}
\usepackage{multirow}
\usepackage{url}
\usepackage{subcaption}

\usepackage{amsmath, amssymb}
\usepackage{algorithmic}

\usepackage[export]{adjustbox}
\usepackage{xcolor} 

\definecolor{lightgreen}{RGB}{188, 245, 188}
\definecolor{lightred}{RGB}{255, 214, 215}

\usepackage{hyperref}

\newcounter{RubingCommentCounter}
 \journal{Information Sciences}

\begin{document}

\begin{frontmatter}

\title{Short-term electricity load forecasting with multi-frequency reconstruction diffusion}

\author[MA1]{Qi Dong}
\ead{3230006098@student.must.edu.mo}
\affiliation[MA1]{organization={School of Computer Science and Engineering, Macau University of Science and Technology},
           city={Macau SAR},
           postcode={999078},
           country={China}}

\author[MA1,MA2]{Rubing Huang\corref{cor1}}
\ead{rbhuang@must.edu.mo}
\affiliation[MA2]{organization={Macau University of Science and Technology Zhuhai MUST Science and Technology Research Institute},
           city={Zhuhai},
           state = {Guangdong},
           postcode={519099},
           country={China}}

\author[MA1]{Ling Zhou}
\ead{lzhou@must.edu.mo}

\author[NT]{Dave Towey}
\ead{dave.towey@nottingham.edu.cn}
\affiliation[NT]{organization={School of Computer Science, University of Nottingham Ningbo China}, 
            city = {Ningbo}, 
            state = {Zhejiang},
            postcode = {315100},
            country = {China}}

\author[MA1]{Jinyu Tian}
\ead{jytian@must.edu.mo}

\author[MA3]{Jianzhou Wang}
\ead{jzwang@must.edu.mo}
\affiliation[MA3]{organization={Department of Engineering Science, Macau University of Science and Technology},
           city={Macau SAR},
           postcode={999078},
           country={China}}

 \cortext[cor1]{Corresponding author.}

\begin{abstract}
    Diffusion models have emerged as a powerful method in various applications.
    However, their application to \textit{Short-Term Electricity Load Forecasting} (STELF)  — a typical scenario in energy systems — remains largely unexplored.
    Considering the nonlinear and fluctuating characteristics of the load data, effectively utilizing the powerful modeling capabilities of diffusion models to enhance STELF accuracy remains a challenge.
    This paper proposes a novel diffusion model with multi-frequency reconstruction for STELF, referred to as the \textit{Multi-Frequency-Reconstruction-based Diffusion} (MFRD) model.
    The MFRD model achieves accurate load forecasting through four key steps: 
    (1) The original data is combined with the decomposed multi-frequency modes to form a new data representation; 
    (2) The diffusion model adds noise to the new data, effectively reducing and weakening the noise in the original data; 
    (3) The reverse process adopts a denoising network that combines \textit{Long Short-Term Memory} (LSTM) and Transformer to enhance noise removal; and 
    (4) The inference process generates the final predictions based on the trained denoising network.
    To validate the effectiveness of the MFRD model, we conducted experiments on two data platforms: \textit{Australian Energy Market Operator} (AEMO) and \textit{Independent System Operator of New England} (ISO-NE).
    The experimental results show that our model consistently outperforms the compared models.
\end{abstract}


\end{frontmatter}

\section{Introduction}

\textit{Short-Term Electricity Load Forecasting} (STELF) typically refers to predictions within an hour, a day, or a week.
It is usually applied to short-term scheduling and decision-making.
STELF is a key component in driving the evolution of future smart grids, helping to advance their construction~\citep{wu2021online}.
In the future, load-forecasting systems will be a vital tool for power engineering professionals.
As smart grids continue to evolve, power systems are demonstrating increasingly higher levels of intelligence and sustainability.
STELF, as an important technology supporting this transformation, is also attracting increasing amounts of attention \citep{fan2024hybrid,javed2022novel,wang2023multitask,wang2025stochastic,yang2025hybrid}.

External factors 
---
such as the increasing penetration of renewables, shifts in economic and social behavior, and rapid urban expansion
---
contribute to the significant complexity and high volatility of electricity data, making STELF a challenging task~\citep{xiao2023short}.
Most of the literature on STELF incorporates some external factors (such as weather conditions, seasonal characteristics, and socio-economic variables)~\citep{haque2022short,wang2023short}. 
These external features enhance training and predictive performance, but increase model complexity.
This can require more computational resources, longer training times, and higher hardware demands~\citep{dong2025building}. 
The complexity can also pose challenges for their implementation.
Furthermore, obtaining multivariate data may be impractical, due to real-world constraints~\citep{dong2025short}. 
Privacy concerns and difficulties accessing relevant data may mean that access to economic or weather-related information is not possible~\citep{tian2024cnns}.
Excessive reliance on external factors can also affect the model's generalization ability~\citep{wang2024reinforcement}. 
Therefore, achieving accurate predictions with lightweight models using only electricity load data remains challenging.
The objective of our model design is to reduce reliance on external data while enhancing its applicability in diverse and resource-constrained environments.

Three main types of techniques have been used for STELF: 
statistical methods, machine learning methods, and \textit{Deep Learning} (DL) methods~\citep{dudek2016neural}.
Most statistical models (such as \textit{AutoRegressive Integrated Moving Average} (ARIMA) and \textit{Seasonal AutoRegressive Integrated Moving Average} (SARIMA)) rely on linear assumptions and are effective for capturing trend, seasonality, and autocorrelation in time-series data~\citep{zhang2023improved}. 
However, real-world load data is shaped by external conditions and characterized by nonlinear complexity:
This limits the effectiveness of linear models.
Although traditional machine learning methods can capture some nonlinearity and high-dimensional features, they may struggle with highly complex nonlinear relationships and intricate feature interactions~\citep{wan2023short}.
Moreover, their performance heavily depends on the data quality, and issues (such as noise, missing values, or anomalies) can significantly degrade predictive accuracy~\citep{wen2024highly}. 
DL models are capable of learning complex nonlinear dependencies in load data.
They can also be scaled relatively easily in depth or width to model more sophisticated patterns~\citep{dong2025short}. 
Although DL methods have been widely applied to STELF~\citep{lv2024multi,mughees2021deep,zhang2025bi}, they remain sensitive to noisy data, which often arises from external disturbances or measurement errors.
Sensitivity to noise reduces the model’s robustness and hinders its ability to learn intricate load patterns:
This can result in a decline in the predictive accuracy, especially when data quality is low or when there are strong external disturbances.

Recently, diffusion-based generative models have attracted widespread attention for their ability to generate high-dimensional data with excellent training stability~\citep{li2024transformer}.
\citet{ho2020denoising} proposed the \textit{Denoising Diffusion Probabilistic Model} (DDPM), a generative framework that gradually transforms random noise into realistic data through a multi-step denoising process. 
This approach enables high-quality image generation by learning how to reverse a noising process in a stable and controllable manner.
These models operate by gradually perturbing the original data into noise, and then learning the reverse process to regenerate high-quality samples.
After achieving good results in image generation, the model has been expanded to other domains, including text and time-series data~\citep{huang2024generative}.
STELF can also be viewed as a generative task:
It involves generating unseen future load data.
The diffusion model is trained using a simple mean squared error loss, and does not require the complex adversarial optimization of generative adversarial networks:
This results in more stable training.
During the forward diffusion process, noise is gradually added, causing the data distribution to shift toward a standard normal form. 
In the reverse, denoising phase, the model reconstructs the data structure, enhancing robustness to noise and improving the prediction stability.
We address the issue of inherent noise present in load data by utilizing the noise injection and denoising mechanisms of the diffusion framework. 
By converting the original data into pure Gaussian noise during the forward diffusion process, the model can better alleviate the impact of the inherent data noise.

To address these challenges, we propose a novel diffusion model for STELF that combines the original data with its decomposed multi-frequency modes:
\textit{Multi-Frequency-Reconstruction-based Diffusion} (MFRD).
The proposed MFRD model consists of the following four steps: 
\begin{enumerate}
    \item 
    \textit{Variational Mode Decomposition} (VMD) decomposes the original data into multiple \textit{Intrinsic Mode Functions} (IMFs) at different frequency scales. 
    The decomposed IMFs are then recombined with the original data to construct multi-frequency features.

    \item 
    The diffusion model employs a gradual noise-adding process, allowing for the suppression of noise disturbances in the original data.

    \item 
    The reverse process is a denoising procedure, where noise is progressively removed from the data. 
    The denoising network is built upon a Transformer model, incorporating a residual \textit{Long Short-Term Memory} (LSTM) module to enhance denoising effectiveness.

    \item 
    The final result is obtained through inference, where the trained denoising network progressively removes noise based on the initial random noise.
\end{enumerate}
In summary, the key contributions of this paper are: 
\begin{itemize}
    \item 
    We only use load data as the input variables, eliminating dependence on external data, and enhancing the model’s adaptability across different scenarios. 
    Multiple frequency components are extracted from the original load data and integrated with it to form a richer, multi-frequency feature representation.
    
    \item  
    To address the challenge of insufficient noise resistance in traditional DL models, we introduce the MFRD model. 
    It employs forward noising and reverse denoising mechanisms to learn clean data representations from noisy observations. 
    This enhances the model’s robustness to noise, improving its
    robustness to noise disturbances.

    \item 
    The denoising network is built on a Transformer architecture with residual LSTM connections. 
    This leverages the Transformer’s global modeling capability and LSTM’s ability to capture temporal dependencies to enhance the denoising performance.

    \item 
    The experiments validate the effectiveness of the proposed model. Evaluations conducted on datasets from two data platforms, comparing MFRD with five methods and seven popular approaches, consistently demonstrate its superiority.
   
\end{itemize}

The rest of this paper is organized as follows: 
Section~\ref{SEC:Related work} introduces the related work. 
Section~\ref{SEC:Preliminary} provides an introduction to the background knowledge for the methods used in the model. 
Section~\ref{SEC:Proposed MFRD} describes the proposed method. 
Section~\ref{SEC:Experiment Settings} presents the experimental settings. 
Section~\ref{SEC:Experiment Results and Analysis} presents and analyzes the experimental results.
Section~\ref{SEC:Conclusions} summarizes the paper, and also discusses potential future work.

\section{Related work
\label{SEC:Related work}}

Research on STELF has been growing. 
In this section, we review the previous work and highlight some recent advances in the field.
Data preprocessing is essential, with data reconstruction being a key technique for reducing data complexity and improving model learning efficiency.
\textit{Empirical Mode Decomposition} (EMD) adaptively decomposes nonlinear, non-stationary signals into IMFs and residual components, enabling time-frequency analysis.
\citet{mathew2021emd} employed a sequence-to-sequence model in conjunction with EMD for data preprocessing.
\citet{mounir2023short} proposed an EMD-based forecasting approach that decomposes stochastic electricity load time series into several IMFs. 
Because VMD~\citep{xiao2024hybrid} offers better noise resistance and alleviates the problem of overlapping frequency components encountered in EMD, it has been gaining increasing attention in STELF~\citep{luo2022ensemble,xia2023combined,zang2021residential}. 
\citet{wu2023novel} used wavelet threshold denoising to remove noise from the data and uses VMD to extract trend features from the denoised data.
\citet{wen2024highly} proposed a two-stage decomposition strategy for load sequences, applying complete ensemble EMD with adaptive noise in the first stage, and VMD in the second.
Decomposition approaches contribute to more accurate and stable load forecasts by isolating significant frequency features from the original data~\citep{dong2025power,liu2025enhancing,yang2025hybrid}.
These things motivated us to use VMD to enhance data preprocessing and multi-frequency feature extraction.

DL-based models have been used to explore the complexities of load forecasting, driving significant research in STELF~\citep{eren2024comprehensive}.
DL model design primarily focuses on the effective processing of temporal sequence data, and on capturing spatial dependencies:
To achieve this, specialized architectures model long-term dependencies, learn temporal patterns, and improve the quality of learned features.
Several time-series forecasting models
---
such as the \textit{Gated Recurrent Unit} (GRU)~\citep{li2022short}, LSTM~\citep{lv2021vmd}, and Transformer~\citep{gao2023adaptive}
---
have been applied to load-forecasting tasks.
To be better able to capture spatial and temporal patterns in load data, some studies have used convolution-based models such as \textit{Convolutional Neural Network} (CNN)~\citep{hong2023short}, \textit{Graph Convolutional Network} (GCN)~\citep{wei2023short}, and \textit{Temporal Convolutional Network} (TCN)~\citep{bian2022research}.
As electricity load data grows, single models are often unable to meet forecasting demands:
This has led to the increasing popularity of hybrid models~\citep{dong2025short}. 
These approaches may involve integrating DL models with data-driven methods or traditional machine learning methods, as well as fusing multiple DL models.
\citet{wang2023multitask} proposed a DL framework combining granular information, quantile regression, and multi-dimensional feature extraction to enhance feature extraction in load forecasting. 
\citet{lin2022hybrid} designed two ensemble models to predict different types of residential loads:
One comprises predictors based on support vector machines, backpropagation neural networks, and generalized regression neural networks; while the other consists of multiple bidirectional LSTMs.
\citet{zhang2023general} proposed a multi-task learning model using TCN and LSTM, where the residual convolutional structure was designed to expand the model’s temporal receptive field.

\citet{wang2024diffload} used a diffusion-based sequence-to-sequence model for uncertainty-aware load forecasting.
Apart from their work, however, it appears that no other research has applied diffusion models to specific STELF tasks.
Several recent studies have, however, applied such models to time-series forecasting, providing valuable guidance for this work. 
\citet{rasul2021autoregressive} proposed TimeGrad, which addresses the multivariate probabilistic time-series forecasting problem:
It uses diffusion probabilistic models to establish connections with energy-based methods.
DL methods have also been explored as denoising networks within the diffusion framework for time-series forecasting tasks~\citep{shen2024multi,yuan2024diffusionts}.
Building on this prior work, we explore a multi-frequency reconstruction-based diffusion model in STELF, using its generative capabilities and DL-based denoising networks to improve forecasting accuracy.

\section{Preliminaries
\label{SEC:Preliminary}}

\subsection{Variational mode decomposition}

VMD is a variational optimization-based method for decomposing a complex signal \( f(t) \) into IMFs with narrowband spectra. 
Given a signal \( f(t) \), the goal is to decompose it into \( K \) modes \( u_k(t) \), such that:
\begin{equation}
   f(t) = \sum_{k=1}^{K} u_k(t),
\end{equation}
where  \( u_k(t) \) denotes the \( k \)-th mode, and \( K \) is the total number of modes.

The frequency-domain representation of each mode \( u_k(t) \) is given by:
\begin{equation}
    \hat{u}_k(\omega) = \mathcal{F}\{u_k(t)\},
\end{equation}
where \( \mathcal{F} \) denotes the Fourier transform.

The bandwidth of each mode is approximated by the squared \( L^2 \)-norm of the derivative after frequency shifting:
\begin{equation}
    \left\|\partial_t \left(u_k(t) e^{-j\omega_k t}\right)\right\|_2^2,
\end{equation}
where \( \omega_k \) denotes the center frequency of each mode.

The VMD objective is to minimize the total bandwidth of all modes, while ensuring signal reconstruction:
\begin{equation}
    \min_{\{u_k\}, \{\omega_k\}} \left\{ \sum_{k=1}^{K} \left\|\partial_t \left(u_k(t) e^{-j\omega_k t}\right)\right\|_2^2 \right\}.
\end{equation}

To solve this constrained optimization problem, VMD constructs the following Lagrangian function:
\begin{equation}
\begin{aligned}
    \mathcal{L}(\{u_k\}, \{\omega_k\}, \lambda) 
        = &\ \alpha \sum_{k=1}^{K} \left\|\partial_t \left(u_k(t) e^{-j\omega_k t}\right)\right\|_2^2 
        + \left\|f(t) - \sum_{k=1}^{K} u_k(t)\right\|_2^2 \\ & 
        + \left\langle \lambda(t), f(t) - \sum_{k=1}^{K} u_k(t) \right\rangle,
\end{aligned}
\end{equation}
where \( \alpha \) is the penalty parameter for the bandwidth constraint, and \( \lambda \) is the Lagrange multiplier.

Using the alternating direction method~\citep{ahajjam2022experimental}
of multipliers, the modes \( u_k(t) \) and center frequencies \( \omega_k \) are iteratively optimized until convergence.

\subsection{Diffusion framework}

The diffusion framework~\citep{ho2020denoising}, comprising a forward and a reverse process:
The forward process involves adding noise to a data distribution, while the reverse process removes noise from the noisy data.

Starting with an initial data distribution \( x_0 \sim q(x) \), Gaussian noise is progressively added through a forward diffusion process, resulting in the final distribution \( x_T \sim \mathcal{N}(0, \mathbf{I}) \), where 
\(\mathbf{I}\) 
denotes the identity covariance matrix.
If \( T \) denotes the total number of noise-addition steps, then this transition can be described by the following equation:
\begin{equation}
    q(x_t | x_{t-1}) = \mathcal{N}\left(x_t; \sqrt{1-\beta_t} \, x_{t-1}, \beta_t \mathbf{I}\right), 
\label{eq:transition}
\end{equation}
where \(\beta_t\) denotes a given increasing variance, and satisfies \(\beta_t \in (0, 1)\).
This entire noise-adding process is represented as a Markov chain:
\begin{equation}
    q(x_{1:T} \mid x_0) = \prod_{t=1}^{T} q(x_t \mid x_{t-1}).
\end{equation}
If \(\bar{\alpha}_t = \prod_{s=1}^{t} \alpha_s, \alpha_t = 1 - \beta_t\), then \( x_t \) can be computed as:
\begin{equation}
    x_t = \sqrt{\bar{\alpha}_t} x_0 + \sqrt{1 - \bar{\alpha}_t} \epsilon,
\end{equation}
where \( \epsilon \) is a noise from \( \mathcal{N}(0, \mathbf{I}) \).

The reverse process involves gradually removing noise from the Gaussian noise \( x_T \) obtained in the forward process.
It is also defined as a Markov chain and involves training a denoising network.
This process begins with \( p(x^T) \) following a Gaussian distribution \( \mathcal{N}(x^T; 0, \mathbf{I}) \), with the overall transition process defined as:
\begin{equation}
    p_\theta(x^{0:T}) := p(x^T) \prod_{t=1}^{T} p_\theta(x^{t-1} \mid x^t),
\end{equation}
where \( p_\theta(x^{0:T}) \) denotes the joint probability distribution over all diffusion steps.

The single-step sampling $p_\theta(x_{t-1} \mid x_t)$ can be written as:
\begin{equation}
    p_\theta(x_{t-1} \mid x_t) = \mathcal{N}(x_{t-1}; \mu_\theta(x_t, t), \Sigma_\theta(x_t, t)\mathbf{I}),
\end{equation}
where \(\mu_\theta : \mathbb{R}^D \times \mathbb{N} \to \mathbb{R}^D\) represents the mean, and \(\Sigma_\theta : \mathbb{R}^D \times \mathbb{N} \to \mathbb{R}^+ \) represents the variance.
The two functions take the variable \(x_t \in \mathbb{R}^D \) and t \(\in \mathbb{N}\) as inputs
and share the parameter set \(\theta\).

The prediction process then uses a denoising network \(\epsilon_\theta\) to directly recover \(x_0\) from \(x_t\).
The parameterization of \(p_\theta(x_{t-1} \mid x_t)\) is chosen as~\citep{fan2024mgtsd}:
\begin{equation}
    \mu_\theta(x_t, t) \colon = \frac{1}{\sqrt{\alpha_t}} 
        \left( 
            x_t - \frac{1 - \alpha_t}{\sqrt{1 - \bar{\alpha}_t}} 
            \epsilon_\theta 
            \left(x_t, t\right) 
        \right).
\label{equ:denoising}
\end{equation}

\citet{yuan2024diffusionts} showed that the parameter set \(\theta\) is learned by minimizing the following loss function:
\begin{equation}
    \mathcal{L}(x_0) = \mathbb{E}_{q(x_t \mid x_0)} \left\| \mu(x_t, x_0) - \mu_\theta(x_t, t) \right\|^2,
\label{equ:loss}
\end{equation}
where \(\mu(x_t, x_0)\) and \(\mu_\theta(x_t, t)\) represent the true and predicted means, respectively.

\subsection{Long Short-Term Memory}

LSTM is a specialized \textit{Recurrent Neural Network} (RNN) architecture:
It is designed to capture and process long-term dependencies in sequential data. 
LSTM uses a memory cell and a gating mechanism to capture temporal dependencies and patterns in load data~\citep{haque2022short}. 
LSTM has strong sequential modeling capability, and has been widely applied in STELF, including for load data with nonlinearities, complex fluctuations, and multiple periodic patterns~\citep{das2020occupant}. 
To better capture temporal dependencies, we use an LSTM framework with residual connections.
Residual connections alleviate the vanishing gradient issue in deep networks by enabling more efficient information propagation across layers. 
They also help retain essential features, enhancing the model’s capacity to learn long-term dependencies.

\subsection{Transformer}

The Transformer model, built on self-attention, is popular for time-series forecasting, due to its ability to capture complex features and dependencies~\citep{gao2023adaptive}.
It excels at capturing long-term dependencies and supports parallel computation, making it particularly well-suited for processing large-scale, high-dimensional, and complex data that is dependent.
The Transformer consists of encoder and decoder components:
The encoder extracts abstract representations from the input data, while
the decoder generates prediction results based on the encoder's output.
The Transformer’s self-attention mechanism captures dependencies among tokens at varying positions within the input.
By mapping the input to \textit{Query} (Q), \textit{Key} (K), and \textit{Value} (V) vectors, the attention weights are computed using the following formula:
\begin{equation}
    {Attention}(Q, K, V) = {softmax}\left( \frac{QK^\mathrm{T}}{\sqrt{D_k}} \right)V,
\end{equation}
where \(D_k\) represents the dimensionality of the key vectors. 
This mechanism calculates the similarity between queries and keys, assigning attention weights that are then applied to the value vectors, thereby focusing on important features.
To enhance representation, the Transformer uses multi-head attention, which learns feature dependencies in multiple subspaces, and then aggregates them into a complete output~\citep{vaswani2017attention}:
This method allows the model to learn more comprehensive and varied feature representations.

\begin{figure*}[!b]
  \centering
  \includegraphics[width=1.0\textwidth]{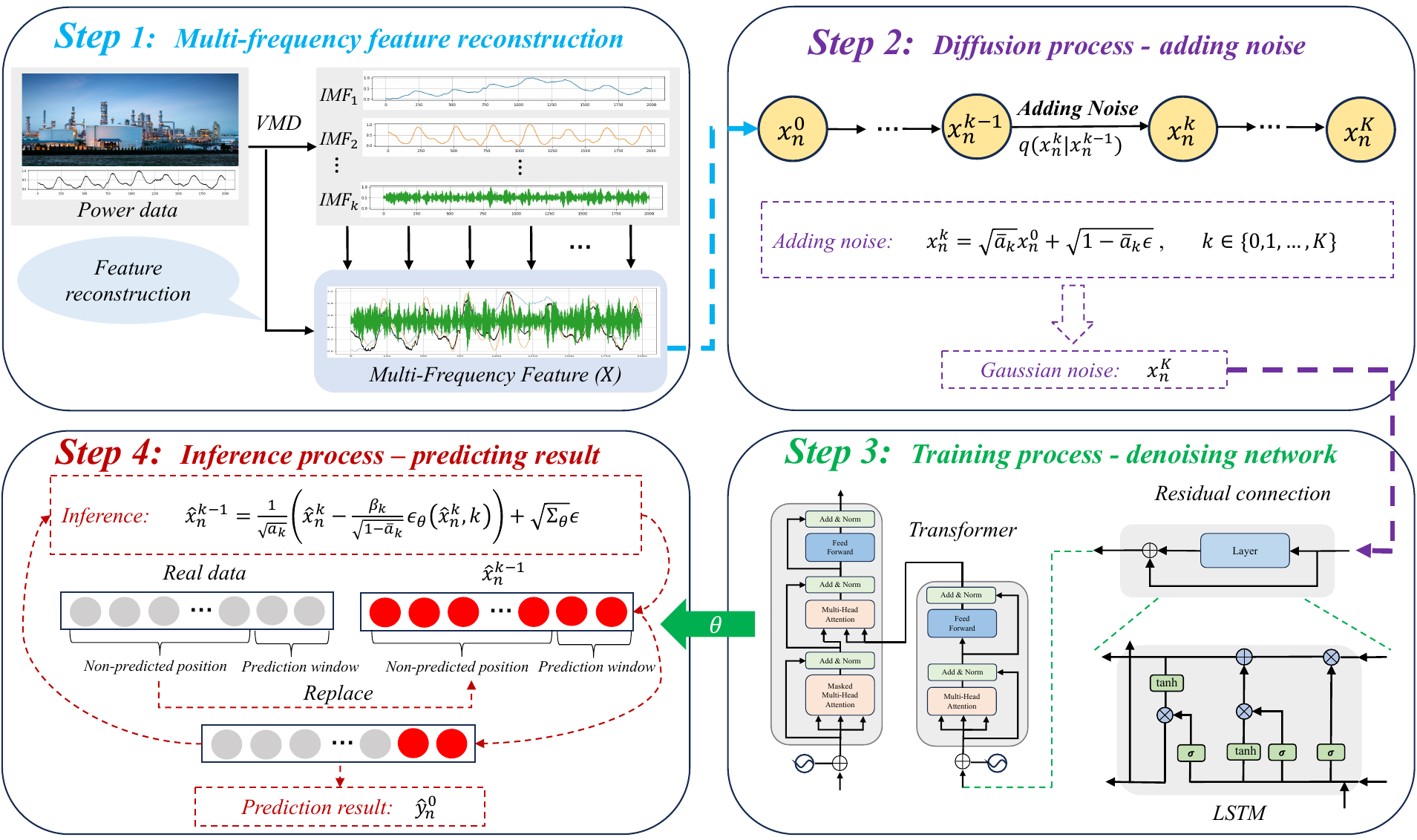 }
  \caption{Overall MFRD framework.}
  \label{fig:framework}
\end{figure*}

\section{Proposed MFRD model
\label{SEC:Proposed MFRD}}

This section provides an overview of the proposed MFRD model.

\subsection{Overall framework}

MFRD constructs a new data representation by combining the original load data with its decomposed multi-frequency modes:
It integrates residual LSTM and Transformer modules within a diffusion framework.
The diffusion framework includes a forward noise-adding process and a reverse denoising process.
The final prediction is obtained through inference.
Figure~\ref{fig:framework} shows how MFRD performs STELF through four main steps: 
the multi-frequency feature reconstruction;
the diffusion process;
the training process; and
the inference process.
MFRD involves mapping the input \( x \) to the output \( Y \). 
The objective of the first three stages is to train an accurate denoising network based on the input \( x \), which is then used to infer the final prediction \( Y \).
This can be concisely formulated as:
\begin{equation}
    Y = \textit{Infer} \longrightarrow 
        \left\{ 
            \mathcal{N}(0, \sigma^2),\; \mathbf{\mathcal{L}_\theta} \left| \hspace{2pt}
            \textit{Diff} \,\Bigl( 
                                \textit{Rec} \bigl[ x, \textit{VMD}(x) \bigr] 
                            \Bigr) 
            \hspace{1pt} \right| 
        \right\},
\end{equation}
where \textit{Rec} refers to the multi-frequency feature reconstruction process; 
\textit{Diff} is the diffusion process; 
\(\mathcal{L}_\theta\) is the target denoising network; 
\(\mathcal{N}(0, \sigma^2)\) represents the Gaussian noise; and 
\textit{Infer} is the inference process.

There are four steps involved in constructing the MFRD:
\begin{itemize}
    \item 
    \textit{Step 1: Multi-frequency feature reconstruction:}
    This step constructs a new data representation by reconstructing the original data together with the multi-frequency modalities. 
    This enables the model to capture richer temporal dynamics, and uncover latent patterns that are difficult to identify from the original data alone.

    \item 
    \textit{Step 2: Diffusion process:} 
    This step gradually corrupts the newly constructed data representation into Gaussian noise. 
    This not only prepares the model for the subsequent denoising, but also reduces the impact of the intrinsic noise present in the original data.

    \item 
    \textit{Step 3: Training process:}
    This involves designing a denoising network based on residual LSTM and Transformer architectures. 
    This denoising network becomes an efficient and accurate model for the subsequent inference phase.

    \item 
    \textit{Step 4: Inference process:} 
    This step iteratively applies the denoising network and generates the final forecasting results. 
\end{itemize}
Each of these four steps will be explained in detail in the following sections.

\subsection{Step 1: Multi-frequency feature reconstruction}
Multi-frequency feature reconstruction decomposes the original data into distinct frequency modes, then combines them with the original input to form a new feature structure.
These multi-frequency modes are introduced as auxiliary features during model training, helping the diffusion model to capture multi-scale temporal dependencies. 
Combined with the original data, they help preserve key information, and improve noise suppression:
This enhances the model's ability to extract patterns from the noisy load data.
The multi-frequency modes of the original data are obtained through VMD, as described in Section~\ref{SEC:Preliminary}.
Since VMD performs a global decomposition over the entire input signal, we apply it independently to each sample.
The original data, denoted \( x_t \), are decomposed into \( k \) distinct frequency modes, represented as \( (v_1, v_2, \dots, v_k) \).
The value of \( k \) typically ranges between $2$ and $10$~\citep{dragomiretskiy2013variational}, 
with the exact value determined based on experimental tuning.
The new data \( X \), obtained after multi-frequency feature reconstruction, can be expressed as: 
\begin{equation}
     X = [x_t, v_1, v_2, \dots, v_k].
\end{equation}

\( X \) retains the original load characteristics while embedding diverse multi-frequency components:
This provides more informative inputs for the modeling.
The reconstructed input serves as the foundation for the diffusion process, where the model progressively denoises the perturbed data, and reconstructs accurate forecasting targets.

\subsection{Step 2: Diffusion process}

The diffusion process injects noise into the data over a sequence of steps.
Unlike the methods by \cite{shen2024multi} and~\cite{li2024transformer}, which apply noise only to the prediction window, our approach introduces noise to the entire sample, encompassing both the look-back and prediction windows~\citep{yuan2024diffusionts}.
For a sample \( x_n = \{x_{n,1}, x_{n,2}, \dots, x_{n,M}\}\) 
---
where
\( n \in \{1, 2, \dots, N\} \),  
\( N \) is the total number of samples, and 
\( M \) denotes the size of each sample
---
the noise-adding process transforms \( x_n^0 \) into \( x_n^K \), as described in Eq. (\ref{eq:transition}), with \( x_n \) initially set to \( x_n^0 \).
We use \( x_n^k \) to represent the data after noise addition at the \( k \)-th step:
\begin{equation}
    x_n^k = \sqrt{\bar{\alpha}_k} x_n^0 + \sqrt{1 - \bar{\alpha}_k} \epsilon, \quad k \in \{0, 1, \dots, K\},
\end{equation}
where \( \epsilon \) is noise from \( \mathcal{N}(0, \mathbf{I}) \).

The progressive noise injection gradually transforms the reconstructed data distribution into a standard Gaussian distribution. 
This forward diffusion process disrupts the intrinsic structure of the data, enabling the model to capture deeper and more enduring characteristics in the reverse process.
This progressive noise injection establishes a structured learning objective for the reverse process.

\subsection{Step 3: Training process}

The training process involves reverse denoising while simultaneously optimizing the neural network.
This process removes noise from the Gaussian noise \( x_n^K \) generated in the forward process, recovering the original data \( x_n^0 \).
This is performed by a specially-designed neural network, as illustrated in Figure~\ref{fig:framework}:
\( x_n^k \) is first processed through a residual LSTM module to capture the temporal dependencies within the sample:
\begin{equation}\label{equ:LSTM}
    H_n^k = \mathcal{LSTM}(x_n^k; \theta),
\end{equation}
\vspace{-15pt}
\begin{equation}\label{equ:residual}
    x_n^k = H_n^k + x_n^k,
\end{equation}
where \( H_n^k = \{h_{n,1}^k, h_{n,2}^k, \dots, h_{n,T}^k\} \) represents the hidden state sequence with a shape of \(T \times \text{hidden\_size}\), \( T \) denotes the window length, and \( \theta \) represents the learned shared parameters.

Both \( x_n^k \) and \( k \) are then passed to the Transformer for denoising, producing an estimate \( \hat{x}_n^0 \):
\begin{equation}\label{equ:Transformer}
    \hat{x}_n^0(x_n^k, k, \theta) = \mathcal{T}(x_n^k + {PosEnc}(k); \theta),
\end{equation}
where \({PosEnc}(k)\) represents the positional encoding of time step \(k\) for incorporating temporal information, and \(\mathcal{T}\) represents the Transformer network.
The parameter \( \theta \) is learned during training by minimizing the loss function, which consists of a base loss function similar to Eq. (\ref{equ:loss}), and a Fourier-based loss function~\citep{fons2022hypertime}. 
This application of Fourier-based loss enhances the capture of frequency-domain features, and reduces the incorrect learning of noise~\citep{yuan2024diffusionts}.
The base and Fourier-based loss functions are:
\begin{equation}
    \mathcal{L}_{base} = \mathbb{E}_{q(x_n^k \mid x_n^0)} \left\| x_n^0 - \hat{x}_n^0(x_n^k, k, \theta) \right\|^2,
\end{equation}
\begin{equation}
    \mathcal{L}_{Fourier} = \mathbb{E}_{q(x_n^k \mid x_n^0)} \left\|\mathcal{F}(x_n^0) - \mathcal{F}(\hat{x}_n^0(x_n^k, k, \theta)) \right\|^2,
\end{equation}
where \(\mathcal{F}\) denotes the Fourier transform.

The training objective \( \mathcal{L}_\theta \) is expressed as:
\begin{equation}\label{equ:ourloss}
    \begin{aligned}
    \mathcal{L}_\theta = \mathbb{E}_{q(x_n^k \mid x_n^0)} 
        \Big[ w_k 
            \Big( 
                & \lambda_1 \|x_n^0 - \hat{x}_n^0(x_n^k, k, \theta)\|^2 + \\
                & \lambda_2 \|\mathcal{F}(x_n^0) - \mathcal{F}(\hat{x}_n^0(x_n^k, k, \theta))\|^2 
            \Big) 
        \Big],
    \end{aligned}
\end{equation}
\begin{equation}
    w_k = \frac{\gamma \alpha_k (1 - \bar{\alpha}_k)}{\beta_k^2},
\end{equation}
where \( \lambda_1 \) and \( \lambda_2 \) are balance weights for the loss terms;
\( w_k \) represents the weight coefficient at each time step \( k \)
(it adjusts the impact of the loss function~\citep{ho2020denoising});
\( \gamma \) is a fixed coefficient set to 0.01~\citep{yuan2024diffusionts}; 
\( \beta_k \) denotes a predefined increasing variance satisfying \( \beta_k \in (0, 1) \); and 
\( \bar{\alpha}_k = \prod_{s=1}^{k} \alpha_s \), with \( \alpha_k = 1 - \beta_k \).

Algorithm~\ref{algorithm:1} presents the training process based on the objective function.

\begin{algorithm}[!t]
\caption{The Training process.}
\label{algorithm:1}
\begin{algorithmic}[1]
    \STATE \textbf{Input:} data \(x_n^k\), initial parameters $\boldsymbol{\theta}$
    \REPEAT   
            \STATE Random $k \sim \text{Uniform}(\{0,1,  \dots, K\})$ and $\boldsymbol{\epsilon} \sim \mathcal{N}(0, \mathbf{I})$
            \STATE Generate diffused sample \(x_n^k\) by 
            \[x_n^k = \sqrt{\bar{\alpha}_k} x_n^0 + \sqrt{1 - \bar{\alpha}_k}\epsilon \] 
            \vspace{-20pt}
            \STATE Obtain \(H_n^k\) using Eq. (\ref{equ:LSTM})
            \STATE Obtain \(x_n^k\) again through a residual connection between \(x_n^k\) and \(H_n^k\) using Eq. (\ref{equ:residual})
            \STATE  Obtain \(\hat{x}_n^0\) by providing input \(x_n^k\) , \(k\) , and \(\boldsymbol{\theta}\) to Eq. (~\ref{equ:Transformer})
            \STATE Calculate the loss $\mathcal{L}_\theta$ in Eq. (\ref{equ:ourloss})
            \STATE Execute a backpropagation based on $\nabla_{\boldsymbol{\theta}} \mathcal{L}(\boldsymbol{\theta})$

    \UNTIL{converged}
\end{algorithmic}
\end{algorithm}

\subsection{Step 4: Inference process}

\begin{algorithm}[!t]
\caption{The inference process.}
\label{algorithm:2}
\begin{algorithmic}[1]
\STATE \textbf{Input:} random noise \(\hat{x}_n^K \sim \mathcal{N}(0, \mathbf{I})\)
\FOR{$k = K~to~1 $}
    \IF{$k > 1$}
    \STATE \( \epsilon \sim \mathcal{N}(0, \mathbf{I}) \)
    \ELSE
    \STATE $\epsilon = 0 $  
    \ENDIF
    \STATE Calculate \(\hat{x}_n^{k-1} \) from \(\hat{x}_n^{k} \) by
    \vspace{-10pt}
    \[\hat{x}_n^{k-1} = \frac{1}{\sqrt{\alpha_k}} 
      \left( \hat{x}_n^k - \frac{\beta_k}{\sqrt{1 - \bar{\alpha}_k}} 
      \boldsymbol{\epsilon}_\theta(\hat{x}_n^k, k) \right) + \sqrt{\Sigma_\theta} \epsilon\]
      \vspace{-10pt}
     \STATE Obtain \(\hat{x}_n^{k-1} \) again by using real data to replace the non-predicted positions   
\ENDFOR
     \STATE Obtain \( \hat{y}_n^0 \) by using the predicted position data from sample \( \hat{x}_n^0 \)
\RETURN \( \hat{y}_n^0 \)
\end{algorithmic}
\end{algorithm}

The detailed inference process is presented in Algorithm~\ref{algorithm:2}.
Initially, some random noise \( \hat{x}_n^K \sim \mathcal{N}(0, \mathbf{I}) \) is added.
Based on Eq. (\ref{equ:denoising}), an intermediate denoising step from \( \hat{x}_n^k \) to \( \hat{x}_n^{k-1} \) can be written as:
\begin{equation}
    \hat{x}_n^{k-1} = \frac{1}{\sqrt{\alpha_k}} 
        \left( 
            \hat{x}_n^k - \frac{\beta_k}{\sqrt{1 - \bar{\alpha}_k}} 
            \boldsymbol{\epsilon}_\theta(\hat{x}_n^k, k) 
        \right) 
        + \sqrt{\Sigma_\theta} \epsilon,
\end{equation}
where \(\Sigma_\theta : \mathbb{R}^D \times \mathbb{N} \to \mathbb{R}^+\) represents the variance, and \(\epsilon_\theta\) denotes the denoising network. 
The term \( \epsilon \) falls in the range 
\( \mathcal{N}(0, \mathbf{I}) \), when \( k > 1 \); 
but is \( 0 \), otherwise.

Because noise is added to the entire sample, true values are used to replace the non-predicted positions after each intermediate denoising step.
This guides the model towards generating prediction results: 
The predicted value \( \hat{x}_n^0 \) represents the sample-level 
prediction scale. 
If the prediction length is \( L \), then the last \( L \) time steps of \( \hat{x}_n^0 \) are extracted as \( \hat{y}_n^0 \), serving as the final desired \( L \)-length prediction result.

\section{Experimental Settings
\label{SEC:Experiment Settings}}

This section discusses the experimental setting details, including the dataset selection, baseline models, parameter settings, and evaluation metrics.

\subsection{Datasets}

This study used datasets from the \textit{Australian Energy Market Operator} (AEMO) for \textit{New South Wales} (NSW), \textit{Queensland} (QLD), and \textit{Victoria} (VIC), as well as the \textit{Independent System Operator of New England} (ISO-NE) in the United States.
The datasets were used to evaluate the proposed model's performance across different regions, time spans, and sampling frequencies. 
The AEMO dataset comprises electricity load data collected from January 1, 2023, to October 31, 2024, with a sampling frequency of 5 minutes.
The ISO-NE dataset comprises hourly electricity load data from January 1, 2003, to December 31, 2014.
These datasets provide a reliable basis for evaluating the performance of our proposed model in terms of accuracy and generalization.

The AEMO datasets from NSW, QLD, and VIC were used for model training and evaluation:
They were divided in an 8:1:1 ratio into a training set, validation set, and test set.
To assess the generalization ability of the model, we used two different splits of the ISO-NE dataset~\citep{dong2023parallel}:
The first group used data from March 1, 2003, to December 31, 2005, for training, with the last 10\% of the set reserved for validation, and data from 2006 serving as the test set; while the second group used data from January 1, 2004, to December 31, 2009, for training, with the last 10\% again reserved for validation, and data from 2010 and 2011 used separately as test sets.

Min-Max normalization was used to scale the data, transforming values to the $[0, 1]$ range.
This standardized feature ranges from avoiding dominance by variables with larger magnitudes during model training.
The normalization formula was:
\begin{equation}
    x' = \frac{x - \min(X)}{\max(X) - \min(X)},
\end{equation}
where \(x\) is the original data value, \(x'\) is the normalized value, \(X\) is the dataset, and \(\min(X)\) and \(\max(X)\) represent the minimum and maximum values of \(X\), respectively.

\subsection{Baseline methods}

We conducted comparative experiments using two different data sources: AEMO and ISO-NE.
MFRD was compared with five representative models on the AEMO dataset: 
ARIMA, RNN, GRU, LSTM, and Transformer.
These models represent a range of widely adopted architectures in load forecasting, allowing for a comprehensive comparison of their predictive capabilities.
MFRD was also compared, using the same parameters and settings, with several other models on the ISO-NE dataset:
SIWNN~\citep{chen2009short};
the original \citep{yu2014incremental} and modified~\citep{cecati2015novel} RBFN-ErrCorr;
WT-ELM-MABC~\citep{li2015short}; 
WT-ELM-PLSR~\citep{li2016ensemble};
ResNetPlus~\citep{chen2018short}; and 
CBR~\citep{dong2023parallel}.
These methods were selected based on their established effectiveness in prior studies and their relevance to electricity load forecasting. 
These comparisons enable a comprehensive evaluation of the proposed model’s predictive performance and adaptability across diverse datasets.

\subsection{Parameter settings}

During the diffusion process, the MFRD model 
introduces noise into the data over 500 steps, and then gradually removes the noise over 200 steps.
This should restore the clear time-series signal. 
The denoising network employs a Transformer-based architecture, comprising a 3-layer encoder and a 2-layer decoder.
We set the initial learning rate to 1.0e-5, and adopted a training strategy with a batch size of 128.
Given the AEMO dataset’s 5-minute sampling frequency, the sample sequence length was set to 288, which corresponds to the number of data points in 24 hours.
Based on this, the model generated 12 data points, corresponding to a 1-hour prediction horizon.
Using the ISO-NE dataset with an hourly sampling frequency, the model was designed with a sequence length of 24 and generated a 1-hour-ahead forecast.
Experiments were set up in Python 3, on a system with an NVIDIA GeForce RTX 4090 GPU, ensuring sufficient computational resources for model training and evaluation.

\subsection{Evaluation metrics
\label{SEC:Evaluation Metrics}}

The performance of the proposed model was examined using several assessment criteria and metrics, including:
\textit{Root Mean Square Error} (RMSE);
\textit{Mean Absolute Error} (MAE);
\textit{Mean Absolute Percentage Error} (MAPE); and 
\textit{coefficient of determination} ($R^2$).  
The formulas for RMSE, MAE, MAPE, and $R^2$ are:
\begin{equation}
    \text{RMSE} = \sqrt{\frac{1}{N} \sum_{i=1}^N (\hat{y}_i - y_i)^2},
\end{equation}
\begin{equation}
    \text{MAE} = \frac{1}{N} \sum_{i=1}^N |\hat{y}_i - y_i|,
\end{equation}
\begin{equation}
    \text{MAPE} = \frac{1}{N} \sum_{i=1}^N \left| \frac{\hat{y}_i - y_i}{y_i} \right| \times 100\%,
\end{equation}
\begin{equation}
    R^2 = 1 - \frac{\sum_{i=1}^N (y_i - \hat{y}_i)^2}{\sum_{i=1}^N (y_i - \bar{y})^2}.
\end{equation}

\section{Experimental Results and Analysis
\label{SEC:Experiment Results and Analysis}}

This section presents an analysis of the experimental results.

\subsection{Performance on AEMO dataset}

\subsubsection{Performance under different \(k\) in VMD}

\begin{figure*}[!b]
  \centering
  \subfloat[MAE]{%
    \includegraphics[width=0.49\textwidth]{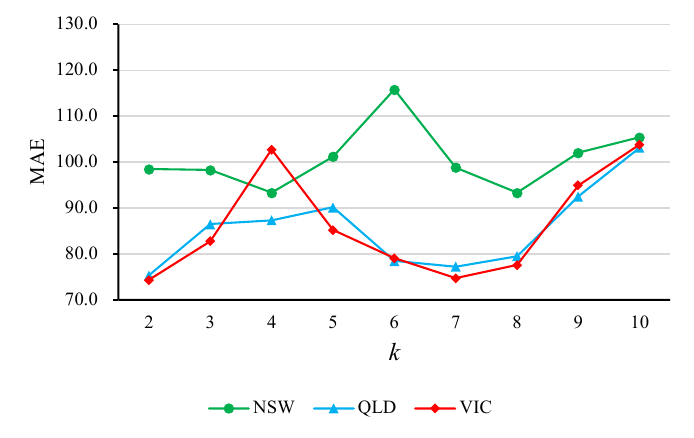}
  }
  \hfill
  \subfloat[RMSE]{%
    \includegraphics[width=0.49\textwidth]{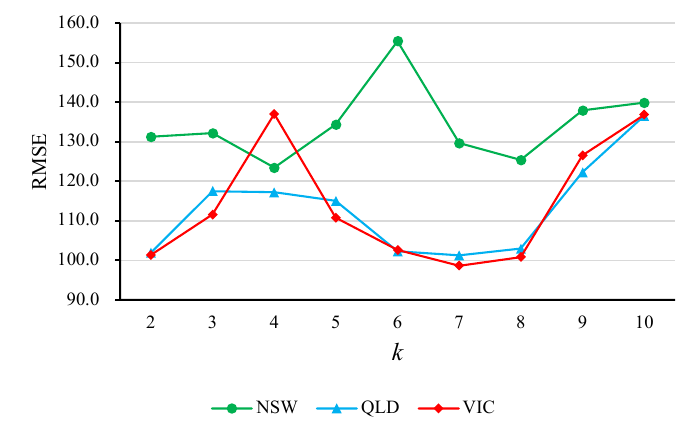}
  }
  
  \subfloat[MAPE]{%
    \includegraphics[width=0.49\textwidth]{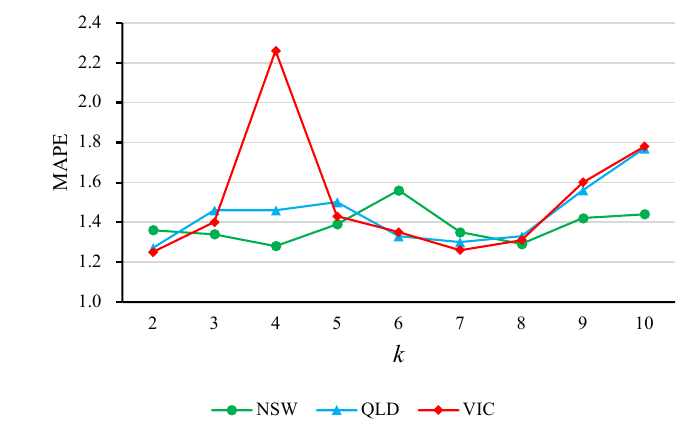}
  }
  \hfill
  \subfloat[$R^2$]{%
    \includegraphics[width=0.49\textwidth]{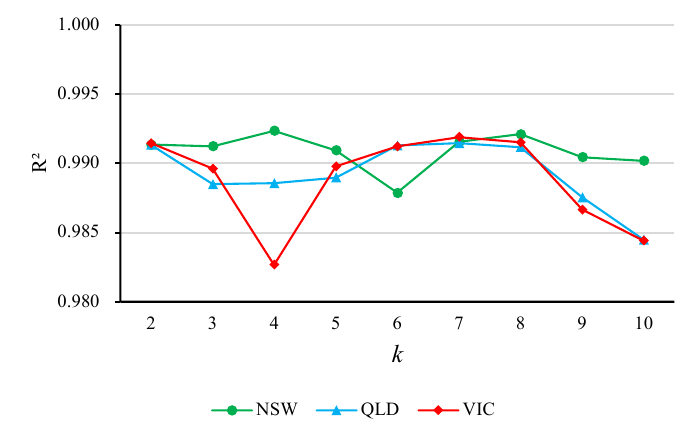}
  }
  
  \caption{Predictive errors for different $k$ values: (a) MAE, (b) RMSE, (c) MAPE, and (d) $R^2$.}
  \label{fig:Different_k}
\end{figure*}

To thoroughly investigate the impact of different frequency modalities, we conducted experiments on the performance of different \(k\) values in VMD, using the NSW, QLD, and VIC datasets.
Different numbers (represented by \(k\)) ranging from 2 to 10 were examined, aiming to identify the optimal amount. 
Figure~\ref{fig:Different_k} shows that the four evaluation metrics (RMSE, MAE, MAPE, and \( R^2 \)) have different trends and significant differences across the three datasets as the value of \(k\) changes. 
This highlights the importance of setting \(k\) appropriately:
It is essential to carefully adjust the \(k\) value based on the characteristics of each dataset and the model's specific performance.

\begin{table*}[!t]
\centering
\scriptsize
\caption{Impact of different VMD mode numbers ($k$) on the performance for NSW, QLD, and VIC datasets. (Values highlighted in \colorbox{lightgreen}{lightgreen} and \colorbox{lightred}{lightred} are the best and second best.)}
\label{tab:performance_different_k}
\renewcommand{\arraystretch}{1.2}
\setlength{\tabcolsep}{2.3mm}{  
\begin{tabular}{c|cccc|cccc|cccc} 
\hline
\multirow{2}*{\textbf{\boldmath$k$}} & \multicolumn{4}{c|}{\textbf{NSW}} & \multicolumn{4}{c|}{\textbf{QLD}} & \multicolumn{4}{c}{\textbf{VIC}} \\
\cline{2-13}
 & MAE & RMSE & MAPE & $R^2$ & MAE & RMSE & MAPE &$R^2$ & MAE & RMSE & MAPE &$R^2$ \\
\hline
2  & 98.49 & 131.20 & 1.36 & \colorbox{lightred}{0.991} & \colorbox{lightgreen}{75.37} & \colorbox{lightred}{101.96} & \colorbox{lightgreen}{1.27} & \colorbox{lightgreen}{0.991} & \colorbox{lightgreen}{74.37} & 101.36 & \colorbox{lightgreen}{1.25} & \colorbox{lightred}{0.991} \\
3  & 98.31 & 132.13 & 1.34 & \colorbox{lightred}{0.991} & 86.53 & 117.51 & 1.46 & 0.988 & 82.85 & 111.61 & 1.40 & 0.990 \\
4  & \colorbox{lightgreen}{93.38} & \colorbox{lightgreen}{123.41} & \colorbox{lightgreen}{1.28} & \colorbox{lightgreen}{0.992} & 87.38 & 117.20 & 1.46 & \colorbox{lightred}{0.989} & 102.75 & 136.97 & 2.26 & 0.983 \\
5  & 101.22 & 134.31 & 1.39 & \colorbox{lightred}{0.991} & 90.23 & 115.03 & 1.50 & \colorbox{lightred}{0.989} & 85.23 & 110.78 & 1.43 & 0.990 \\
6  & 115.84 & 155.38 & 1.56 & 0.988 & 78.52 & 102.27 & 1.33 & \colorbox{lightgreen}{0.991} & 79.10 & 102.65 & 1.35 & \colorbox{lightred}{0.991} \\
7  & 98.88 & 129.64 & 1.35 & \colorbox{lightgreen}{0.992} & \colorbox{lightred}{77.29} & \colorbox{lightgreen}{101.28} & \colorbox{lightred}{1.30} & \colorbox{lightgreen}{0.991} & \colorbox{lightred}{74.78} & \colorbox{lightgreen}{98.69} & \colorbox{lightred}{1.26} & \colorbox{lightgreen}{0.992} \\
8  & \colorbox{lightred}{93.39} & \colorbox{lightred}{125.36} & \colorbox{lightred}{1.29} & \colorbox{lightgreen}{0.992} & 79.55 & 102.97 & 1.33 & \colorbox{lightgreen}{0.991} & 77.61 & 100.88 & 1.31 & \colorbox{lightgreen}{0.992} \\
9  & 102.07 & 137.90 & 1.42 & 0.990 & 92.56 & 122.28 & 1.56 & 0.988 & 95.02 & 126.54 & 1.60 & 0.987 \\
10 & 105.42 & 139.83 & 1.44 & 0.990 & 103.20 & 136.49 & 1.77 & 0.984 & 103.86 & 136.78 & 1.78 & 0.984 \\
\hline
\end{tabular}
}
\end{table*}

Table~\ref{tab:performance_different_k} presents the model's performance on the NSW, QLD, and VIC datasets across different values of \(k\):
There is considerable variation.
The best predictive performance with NSW is when $k = 4$ (with an MAE of $93.38$, RMSE of $123.41$, MAPE of $1.28$\%, and \( R^2 \) of $0.992$), indicating the most accurate fit at this point.
For QLD and VIC, \( k = 2 \) and \( k = 7 \) yield relatively optimal results, respectively, ranking either best or second-best across the evaluation metrics:
QLD with \( k = 2 \) results in an MAE of $75.37$, RMSE of $101.96$, MAPE of $1.27$\%, and \( R^2 \) of $0.991$; and
VIC \( k = 7 \) results in an MAE of $74.78$, RMSE of $98.69$, MAPE of $1.26$\%, and \( R^2 \) of $0.992$.
Overall, the optimal number of modes ($k$) is not uniform across the datasets, but depends on the specific characteristics of each dataset. 
The $R^2$ values for all datasets and $k$ values are close to $1$, indicating a high degree of fit, and validating the effectiveness of the model for load forecasting.

In summary, the findings highlight the need to determine the optimal number of frequency components to ensure predictive effectiveness.
Adjusting $k$, which determines how the frequency components are combined, allows the model to handle varying data patterns flexibly.
This helps it capture underlying patterns and improve the forecasting accuracy more effectively.

\subsubsection{Performance comparison and analysis}

\begin{table*}[!b]
\centering
\scriptsize
\caption{Predictive performance on the NSW, QLD, and VIC datasets. 
(Values highlighted in \colorbox{lightgreen}{lightgreen} and \colorbox{lightred}{lightred} represent the best and second best.)}
\label{tab:performance_comparison}
\renewcommand{\arraystretch}{1.2}
\setlength{\tabcolsep}{1.35mm}{ 
\begin{tabular}{l|cccc|cccc|cccc} 
\hline
\multirow{2}*{\textbf{Approach}} & \multicolumn{4}{c|}{\textbf{NSW}} & \multicolumn{4}{c|}{\textbf{QLD}} & \multicolumn{4}{c}{\textbf{VIC}} \\
\cline{2-13}
 & MAE & RMSE & MAPE & $R^2$ & MAE & RMSE & MAPE &$R^2$ & MAE & RMSE & MAPE &$R^2$ \\
\hline
ARIMA & 153.72 & 219.16 & 2.07 & 0.976 & 129.21 & 183.79 & 2.15 & 0.972 & 131.05 & 183.98 & 2.84 & 0.969 \\
RNN & 213.13 & 278.62 & 2.85 & 0.961 & 123.94 & 172.04 & 2.08 & 0.975 & 155.49 & 206.08 & 3.39 & 0.961 \\
LSTM & 136.72 & 198.14 & 1.82 & 0.980 & 121.69 & 172.32 & 2.06 & 0.975 & 121.43 & 173.05 & 2.59 & 0.972 \\
GRU & \colorbox{lightred}{107.96} & \colorbox{lightred}{145.93} & 1.82 & 0.982 & \colorbox{lightred}{85.26} & \colorbox{lightred}{117.61} & \colorbox{lightred}{1.44} & \colorbox{lightred}{0.988} & \colorbox{lightred}{89.98} & \colorbox{lightred}{126.78} & \colorbox{lightred}{1.96} & \colorbox{lightred}{0.985} \\
Transformer & 109.24 & 151.86 & \colorbox{lightred}{1.51} & \colorbox{lightred}{0.988} & 90.58 & 128.67 & 1.49 & 0.986 & 129.80 & 176.60 & 2.76 & 0.971 \\
MFRD (our method) & \colorbox{lightgreen}{93.38} & \colorbox{lightgreen}{123.41} & \colorbox{lightgreen}{1.28} & \colorbox{lightgreen}{0.992} & \colorbox{lightgreen}{75.37} & \colorbox{lightgreen}{101.96} & \colorbox{lightgreen}{1.27} & \colorbox{lightgreen}{0.991} & \colorbox{lightgreen}{74.78} & \colorbox{lightgreen}{98.69} & \colorbox{lightgreen}{1.26} & \colorbox{lightgreen}{0.992} \\
\hline
\end{tabular}
}
\end{table*}

\begin{figure*}[!b]
  \centering
  \subfloat[MAE]{%
    \includegraphics[width=0.49\textwidth]{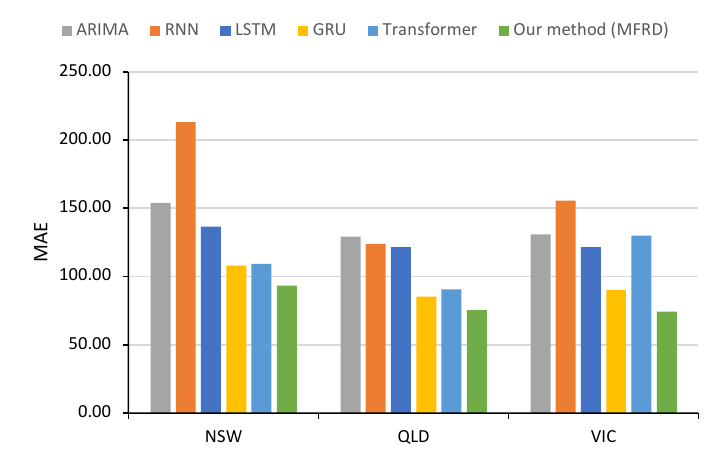}
  }
  \subfloat[RMSE]{%
    \includegraphics[width=0.49\textwidth]{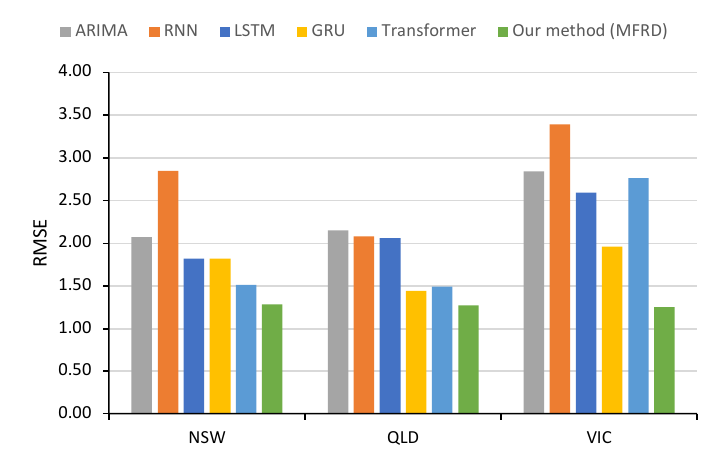}
  }  
  
  \subfloat[MAPE]{%
    \includegraphics[width=0.49\textwidth]{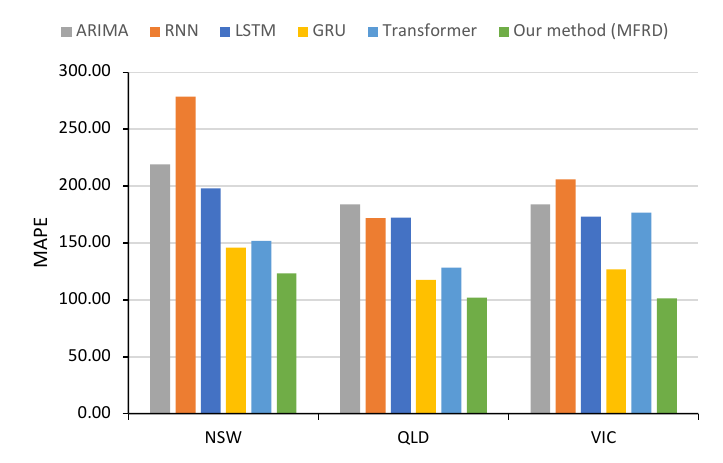}
  }
  \subfloat[$R^2$]{%
    \includegraphics[width=0.49\textwidth]{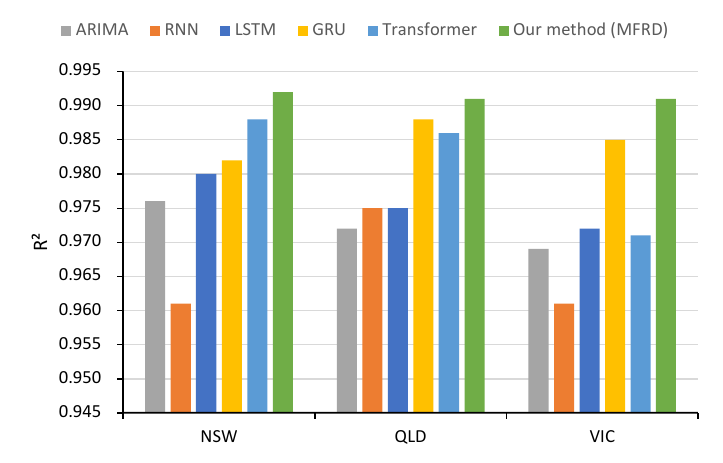}
  }

  \caption{Predictive errors with the NSW, QLD, and VIC datasets: (a) MAE, (b) RMSE, (c) MAPE, and (d) $R^2$. 
}
  \label{fig:comparison of predictive errors AEMO}
\end{figure*}

We compared the performance of MFRD with that of several commonly used models (ARIMA, RNN, LSTM, GRU, and Transformer). 
Table~\ref{tab:performance_comparison} presents a comparative evaluation of the predictive performance across the three datasets (NSW, QLD, and VIC) using the four metrics (MAE, RMSE, MAPE, and $R^2$).
Figure~\ref{fig:comparison of predictive errors AEMO} presents a graphical representation of these results, highlighting the performance improvement of our model over the baselines:
The bar charts clearly show that MFRD (our method) consistently achieves the fewest errors and the highest $R^2$ scores across all datasets. 
The traditional models, ARIMA and RNN, perform relatively poorly across all datasets, typically producing higher error values and lower $R^2$ scores. 
Although the LSTM and Transformer models offer noticeable improvements over the traditional methods, their performance is inconsistent across different regions. 
GRU achieves relatively stable results across datasets, but its overall accuracy lags behind that of MFRD.
With the NSW dataset, for example, MFRD has an MAE of $93.38$, RMSE of $123.41$, MAPE of $1.28$\%, and $R^2$ of $0.992$, significantly outperforming the second-best model (GRU or Transformer). 
The MFRD forecasting results with QLD and VIC are very consistent, reflecting similar levels of load fluctuation and highlighting the model's robustness in handling such temporal characteristics.
With QLD, MFRD achieves an MAE of $75.37$, RMSE of $101.96$, and MAPE of $1.27$\%; and 
with VIC, it achieves an MAE of $74.78$, RMSE of $98.69$, and MAPE of $1.26$\%. 
In both cases, the $R^2$ exceeds $0.990$, making it the highest among all compared methods.

These findings demonstrate that the proposed MFRD model not only captures complex temporal dependencies more effectively, but also maintains consistent performance across regions with varying load characteristics.
The combination of low error rates and high determination coefficients, across all datasets, confirms the robustness and adaptability of MFRD for STELF.

\subsubsection{Ablation studies}

\begin{table*}[!b]
\centering
\scriptsize
\caption{Ablation study results (across the NSW, QLD, and VIC datasets). (Values highlighted in \colorbox{lightgreen}{lightgreen} and \colorbox{lightred}{lightred} represent the best and second best performances.)}
\label{tab:Ablation study}
\renewcommand{\arraystretch}{1.2}
\setlength{\tabcolsep}{1.4mm}{ 
\begin{tabular}{l|cccc|cccc|cccc} 
\hline
\multirow{2}*{\textbf{Approach}} & \multicolumn{4}{c|}{\textbf{NSW}} & \multicolumn{4}{c|}{\textbf{QLD}} & \multicolumn{4}{c}{\textbf{VIC}} \\
\cline{2-13}
 & MAE & RMSE & MAPE & $R^2$ & MAE & RMSE & MAPE &$R^2$ & MAE & RMSE & MAPE &$R^2$ \\
\hline
w/o-VMD & 157.43 & 222.50 & 2.14 & 0.975 & 127.58 & 185.48 & 2.15 & 0.971 & 147.32 & 234.76 & 3.20 & 0.949 \\
w/o-LSTM & \colorbox{lightgreen}{92.33} & \colorbox{lightgreen}{122.76} & \colorbox{lightgreen}{1.27} & \colorbox{lightgreen}{0.992} & \colorbox{lightred}{77.53} & \colorbox{lightred}{105.64} & \colorbox{lightred}{1.30} & \colorbox{lightgreen}{0.991} & \colorbox{lightred}{76.41} & \colorbox{lightred}{103.75} & \colorbox{lightred}{1.28} & \colorbox{lightred}{0.991} \\
w/o-VMD-LSTM & 158.26 & 241.50 & 2.16 & 0.971 & 133.21 & 212.25 & 2.22 & 0.962 & 142.61 & 219.57 & 3.07 & 0.955 \\
w/o-F & 98.93 & 131.74 & 1.35 & \colorbox{lightred}{0.991} & 80.04 & 112.12 & 1.34 & \colorbox{lightred}{0.990} & 78.49 & 107.81 & 1.32 & 0.990 \\
MFRD (our method) & \colorbox{lightred}{93.38} & \colorbox{lightred}{123.41} & \colorbox{lightred}{1.28} & \colorbox{lightgreen}{0.992} & \colorbox{lightgreen}{75.37} & \colorbox{lightgreen}{101.96} & \colorbox{lightgreen}{1.27} & \colorbox{lightgreen}{0.991} & \colorbox{lightgreen}{74.78} & \colorbox{lightgreen}{98.69} & \colorbox{lightgreen}{1.26} & \colorbox{lightgreen}{0.992}\\
\hline
\end{tabular}
}
\end{table*}

The ablation studies involved removing the main model modules to evaluate their contributions to performance improvement.
The experiments included the following variants:
without VMD decomposition (w/o-VMD); 
without LSTM (w/o-LSTM); 
with neither VMD nor LSTM (w/o-VMD-LSTM);
without Fourier transform (w/o-F); and 
the MFRD model.
Table~\ref{tab:Ablation study} presents the results of the ablation study, comparing the performance of the various configurations across the three datasets. 
Compared with the complete method, w/o-VMD had a significantly poorer performance across all datasets, with higher MAE, RMSE, and MAPE values, and a lower $R^2$ value:
MAE increased by $68.59$\%, $69.27$\%, and $97$\%,
on the NSW, QLD, and VIC datasets, respectively.
This highlights the significant contribution of the VMD module to the overall improvement in performance.
This finding suggests that multi-frequency feature reconstruction plays a crucial role in extracting key data features and improving the model's predictive accuracy.
Although w/o-LSTM slightly outperforms MFRD on the NSW dataset, it exhibits a noticeable decline on QLD and VIC, as indicated by MAE, RMSE, and MAPE.
This suggests that although the LSTM enhancement is limited, it can still improve the model's overall generalization abilities. 
The w/o-VMD-LSTM model also had poorer performance than MFRD, across all three datasets:
This further highlights the importance of multi-frequency feature reconstruction and the use of LSTM in improving model performance.
The w/o-F model also performed poorly, across all evaluation metrics and all three datasets, although it did outperform w/o-VMD.
This suggests that the Fourier-based loss function can effectively improve model performance.

The ablation studies demonstrate the key roles that multi-frequency feature reconstruction, LSTM, and Fourier-based loss function play in enhancing model performance. 
By integrating all of these technologies, MFRD achieves more accurate and stable predictions across the datasets. 
These findings not only validate the effectiveness of each component, but also offer valuable insights for further optimizing the model's architecture and enhancing its generalization capability in future work.

\subsection{Performance on ISO-NE dataset}

The ISO-NE dataset was used to assess MFRD’s generalization performance. 
Two groups of experiments compared its performance against representative methods reported in the literature.
The experiments first examined MFRD's performance under different multi-frequency and feature-reconstruction settings. 
The best results were then compared against several existing methods.

\subsubsection{Results of the Group 1 experiment}

The training and validation for the Group 1 experiment utilized data from March 1, 2003, to December 31, 2005, with data from 2006 serving as the test set.
The model's monthly forecasting performance throughout 2006 was evaluated under various multi-frequency feature reconstruction settings.
Table~\ref{tab:monthly_mape_k} shows a stable performance by MFRD, with it consistently achieving competitive MAPE results, for all values of \(k\).
The best overall performance was when \(k = 7\), with an average MAPE of 1.15\%. 
Further examination of the monthly breakdown revealed that \( k = 7 \) yields either the best or the second-best MAPE performance in most months, with only three months showing results slightly worse than the second best.
During periods of high load fluctuation (like July and August), the model still maintains robust performance (with MAPE values of 1.45\% and 1.39\%, respectively), indicating that it is also effective at dealing with seasonal fluctuations.

\begin{table*}[!t]
\centering
\scriptsize
\caption{MFRD MAPE (\%) performance on the ISO-NE dataset for each month of 2006, under different $k$ values. 
(Values highlighted in \colorbox{lightgreen}{lightgreen} and \colorbox{lightred}{lightred} represent the best and second best.)}
\label{tab:monthly_mape_k}
\renewcommand{\arraystretch}{1.2}
\setlength{\tabcolsep}{4.7mm}
{ 
\begin{tabular}{c|ccccccccc}
\hline
\multirow{2}{*}{\textbf{Month}} & \multicolumn{9}{c}{\textbf{\boldmath$k$}} \\ 
\cline{2-10}
 &2 &3 &4 &5 &6 &7 &8 &9 &10 \\
\hline
January   & 1.50 & 1.15 & 1.11 & 1.11 & \colorbox{lightred}{1.08} & \colorbox{lightgreen}{1.05} & 1.16 & 1.18 & 1.18 \\
February  & 1.48 & \colorbox{lightred}{1.06} & 1.08 & 1.13 & 1.21 & \colorbox{lightgreen}{1.05} & 1.16 & 1.18 & 1.15 \\
March     & 1.28 & 1.08 & 1.09 & 1.15 & 1.16 & \colorbox{lightgreen}{1.03} & \colorbox{lightred}{1.05} & 1.09 & 1.09 \\
April     & 1.47 & 1.16 & \colorbox{lightgreen}{1.10} & 1.30 & 1.24 & \colorbox{lightred}{1.14} & 1.15 & \colorbox{lightred}{1.14} & 1.19 \\
May       & 1.47 & 1.11 & \colorbox{lightred}{1.05} & 1.14 & 1.14 & 1.08 & \colorbox{lightgreen}{1.04} & 1.07 & \colorbox{lightred}{1.05} \\
June      & 1.74 & 1.35 & 1.29 & 1.25 & 1.36 & \colorbox{lightred}{1.24} & 1.25 & 1.25 & \colorbox{lightgreen}{1.23} \\
July      & 1.76 & 1.54 & 1.51 & 1.46 & 1.59 & \colorbox{lightred}{1.45} & \colorbox{lightgreen}{1.42} & 1.55 & 1.53 \\
August    & 1.81 & 1.52 & 1.43 & 1.47 & 1.50 & 1.39 & \colorbox{lightgreen}{1.35} & \colorbox{lightred}{1.37} & 1.39 \\
September & 1.73 & 1.25 & 1.10 & 1.15 & 1.16 & 1.10 & \colorbox{lightred}{0.99} & \colorbox{lightred}{0.99} & \colorbox{lightgreen}{0.97} \\
October   & 1.60 & 1.30 & 1.26 & 1.29 & 1.33 & \colorbox{lightgreen}{1.19} & \colorbox{lightred}{1.25} & 1.33 & 1.26 \\
November  & 1.47 & 1.10 & 1.18 & 1.08 & \colorbox{lightred}{1.07} & \colorbox{lightgreen}{0.96} & 1.12 & 1.27 & 1.08 \\
December  & 1.58 & \colorbox{lightred}{1.10} & 1.12 & 1.14 & 1.18 & \colorbox{lightgreen}{1.09} & 1.16 & 1.31 & 1.28 \\
\hline
\textbf{\textit{Average}} & 1.57 & 1.23 & 1.19 & 1.22 & 1.25 & \colorbox{lightgreen}{1.15} & \colorbox{lightred}{1.18} & 1.23 & 1.20 \\
\hline
\end{tabular}
}
\end{table*}

\begin{table}[!t]
\centering
\scriptsize
\caption{MAPE (\%) performance on the ISO-NE dataset in 2006. 
(Values highlighted in \colorbox{lightgreen}{lightgreen} and \colorbox{lightred}{lightred} represent the best and second best.)}
\label{tab:mape_group1_isone}
\renewcommand{\arraystretch}{1.2}
\setlength{\tabcolsep}{2.2mm}
\resizebox{\textwidth}{!}{
\begin{tabular}{c|cccccc}
\hline
\multirow{2}{*}{\textbf{Month}} & \textbf{SIWNN} & \textbf{WT-ELM-PLSR} & \textbf{WT-ELM-MABC} & \textbf{ResNetPlus} & \textbf{CBR} & \textbf{Our method}\\
&\textbf{\citep{chen2009short}} & \textbf{\citep{li2016ensemble}} & \textbf{\citep{li2015short}} & \textbf{\citep{chen2018short}} & \textbf{\citep{dong2023parallel}} & \textbf{(MFRD)}\\
\hline
January   & 1.60 & 1.62 & 1.52 & 1.62 & \colorbox{lightred}{1.33} & \colorbox{lightgreen}{1.05} \\
February  & 1.43 & 1.27 & 1.28 & 1.31 & \colorbox{lightgreen}{1.03} & \colorbox{lightred}{1.05} \\
March     & 1.47 & 1.07 & 1.37 & 1.17 & \colorbox{lightgreen}{0.84} & \colorbox{lightred}{1.03} \\
April     & 1.26 & 1.38 & \colorbox{lightgreen}{1.05} & 1.34 & 1.21 & \colorbox{lightred}{1.14} \\
May       & 1.61 & 1.25 & \colorbox{lightred}{1.23} & 1.32 & 1.25 & \colorbox{lightgreen}{1.08} \\
June      & 1.79 & 1.52 & 1.54 & \colorbox{lightred}{1.41} & 1.49 & \colorbox{lightgreen}{1.24} \\
July      & 2.70 & 2.00 & 2.07 & 1.96 & \colorbox{lightred}{1.88} & \colorbox{lightgreen}{1.45} \\
August    & 2.62 & 1.86 & 2.06 & \colorbox{lightred}{1.55} & 1.62 & \colorbox{lightgreen}{1.39} \\
September & 1.48 & 1.45 & 1.41 & 1.40 & \colorbox{lightred}{1.19} & \colorbox{lightgreen}{1.10} \\
October   & 1.38 & \colorbox{lightred}{1.19} & 1.23 & 1.29 & \colorbox{lightgreen}{1.18} & \colorbox{lightred}{1.19} \\
November  & 1.39 & 1.54 & \colorbox{lightred}{1.33} & 1.51 & 1.36 & \colorbox{lightgreen}{0.96} \\
December  & 1.75 & 1.77 & 1.65 & 1.47 & \colorbox{lightred}{1.28} & \colorbox{lightgreen}{1.09} \\
\hline
\textbf{\textit{Average}} & 1.71 & 1.49 & 1.48 & 1.45 & \colorbox{lightred}{1.31} & \colorbox{lightgreen}{1.15} \\
\hline
\end{tabular}}
\end{table}

Based on the identified optimal configuration (\(k = 7\)), we benchmarked MFRD against five representative models (SIWNN~\citep{chen2009short}, WT-ELM-PLSR~\citep{li2016ensemble}, WT-ELM-MABC~\citep{li2015short}, ResNetPlus~\citep{chen2018short}, and CBR~\citep{dong2023parallel}).
Table~\ref{tab:mape_group1_isone} presents the monthly MAPE values for the six forecasting methods on the ISO-NE dataset in 2006. 
Among the approaches, the proposed MFRD model consistently ranks best or second-best across nearly every month, resulting in the lowest annual average error of 1.15\%. 
All models show reduced accuracy in months with high variability (like July and August), reflecting the inherent difficulty of forecasting under sharp seasonal fluctuations.
However, MFRD had consistently fewer errors during these periods:
It achieved MAPE values of $1.45$\% and $1.39$\% in July and August, while other models had significantly more errors (such as SIWNN (2.70\%, 2.62\%) and ResNetPlus (1.96\%, 1.55\%)).
This consistently good performance, especially with highly variable loads, underscores the strong generalization ability of our method.

\subsubsection{Results of the Group 2 experiment}

The training and validation for the Group 2 experiment used data from January 1, 2004, to December 31, 2009, with data from 2010 and 2011 serving as separate test sets. 
The model was evaluated under different frequency settings.
Table~\ref{tab:performance_k_ISO-NE} presents the MFRD MAPE performance for different \(k\) values:  
The best MAPE was with \(k = 8\) for the 2010 dataset ($1.10$\%), and with \(k = 7\) for the 2011 dataset ($1.51$\%). 
Based on a comparison between \(k = 7\) and \(k = 8\), we selected the results for \(k = 8\), which achieved a slightly better performance, with an MAPE of $1.10$\% in 2010 and $1.52$\% in 2011.

\begin{table*}[!t]
\centering
\footnotesize
\caption{MFRD MAPE (\%) performance on the ISO-NE dataset for 2010 and 2011 for different \(k\) values. 
(Values highlighted in \colorbox{lightgreen}{lightgreen} and \colorbox{lightred}{lightred} represent the best and second best.)}
\label{tab:performance_k_ISO-NE}
\renewcommand{\arraystretch}{1.2}
\setlength{\tabcolsep}{5mm}
\resizebox{\textwidth}{!}
{
\begin{tabular}{c|cccccccccc}
\hline
\multirow{2}{*}{\textbf{Year}} & \multicolumn{9}{c}{\textbf{\boldmath$k$}} \\ \cline{2-10}
&2 &3 &4 &5 &6 &7 &8 &9 &10 \\ 
\hline
2010 & 1.41 & 1.29 & 1.21 & \colorbox{lightred}{1.19} & 1.20 & \colorbox{lightred}{1.19} & \colorbox{lightgreen}{1.10} & 1.23 & 1.21 \\ 
2011 & 2.09 & 1.85 & 1.64 & 1.63 & 1.66 & \colorbox{lightgreen}{1.51} & \colorbox{lightred}{1.52} & \colorbox{lightred}{1.52} & 1.56 \\ 
\hline
\end{tabular}}
\end{table*}

We also compared the MFRD MAPE performance with five representative methods, as shown in Table~\ref{tab:performance_models_ISO-NE}.
The compared methods were:
the original \citep{yu2014incremental} and modified~\citep{cecati2015novel} RBFN-ErrCorr;
WT-ELM-PLSR~\citep{li2016ensemble}; ResNetPlus~\citep{chen2018short}; and CBR~\citep{dong2023parallel}.
The results show a clear pattern, with our proposed MFRD achieving the best MAPE in 2010 ($1.10$\%) and second-best results in 2011 ($1.52$\%).
MFRD had the best results in 2010 ($1.10$\%), outperforming the next best, CBR, by $0.18$ percentage points
---
an improvement of $14.1$\%.
MFRD had the second-best results in 2011 ($1.52$\%), closely behind CBR’s 1.46\%.

\begin{table}[!b]
\centering
\footnotesize
\caption{MAPE (\%) performance evaluation on the ISO-NE dataset for years 2010 and 2011. 
(Values highlighted in \colorbox{lightgreen}{lightgreen} and \colorbox{lightred}{lightred} represent the best and second best.)}
\label{tab:performance_models_ISO-NE}
\renewcommand{\arraystretch}{1.2} 
\setlength{\tabcolsep}{8mm}{
\begin{tabular}{l|cc}
\hline
\textbf{Model} & \textbf{2010} & \textbf{2011} \\
\hline
RBFN-ErrCorr original~\citep{yu2014incremental} & 1.80 & 2.02 \\
RBFN-ErrCorr modified~\citep{cecati2015novel} & 1.75 & 1.98 \\
WT-ELM-PLSR~\citep{li2016ensemble} & 1.50 & 1.80 \\
ResNetPlus~\citep{chen2018short} & 1.50 & 1.64 \\
CBR~\citep{dong2023parallel} & \colorbox{lightred}{1.28} & \colorbox{lightgreen}{1.46} \\
MFRD (our method) & \colorbox{lightgreen}{1.10} & \colorbox{lightred}{1.52} \\
\hline
\end{tabular}}
\end{table}

\subsubsection{Generalization capability}

The results from both groups of experiments confirm the effectiveness and generalization ability of the proposed MFRD method. 
Across the various years' settings, MFRD maintains stable and accurate forecasting performance, highlighting its robustness. 
Real-world electricity load data often exhibit strong seasonal variations and can be subject to significant noise, which is challenging for accurate forecasting:
The proposed model's robustness in the face of such issues is, therefore, critical.
Diffusion also enables the model to identify complex structures better and suppress noise effects, leading to more accurate forecasts in STELF.
Overall, the proposed framework shows strong adaptability and reliability, making it well-suited for practical deployment in dynamic forecasting scenarios.

\section{Conclusions and Future Work
\label{SEC:Conclusions}}

This paper has introduced the MFRD model for STELF.
It addresses issues arising from the complex, volatile, and noisy nature of electricity load data. 
We have improved data representation by reconstructing inherent frequency modes, thereby enhancing the perception of multi-scale patterns.
MFRD utilizes a diffusion framework to enhance its modeling of complex temporal patterns and improve the robustness of its prediction. 
Through the progressive addition of noise, the forward process converts data into Gaussian noise, which helps to suppress the inherent noise and fluctuations.
The reverse process then reconstructs the temporal dependencies, effectively recovering essential patterns.
The denoising network is built on a Transformer backbone, utilizing residual LSTM modules to enhance temporal feature extraction and denoising effectiveness.
The final prediction result is obtained from the trained denoising network during the inference process.
MFRD was evaluated on the AEMO dataset, comparing it with five key models.
We also benchmarked it on the ISO-NE dataset against several recent methods.
The experimental results show that our proposed method outperforms all baseline models, demonstrating superior accuracy and robustness across all datasets.

This research has provided a robust solution for STELF and offers valuable insights for future work that applies diffusion models to STELF tasks.
However, there are still several areas in which future research can be further improved.
Firstly, the computational complexity of diffusion models is relatively high:
Optimizing model efficiency (including reducing computational costs) while maintaining predictive accuracy remains an important research direction. 
Secondly, future research will further explore how diffusion models can effectively capture and model spatial relationships.

This study lays the groundwork for advancing the application of diffusion models in STELF and identifies promising directions for further exploration.

\section*{Acknowledgements}

This work was partially supported by the Science and Technology Development Fund of Macau, Macau SAR, under Grant Nos. 0021/2023/RIA1 and 0035/2023/ITP1.

\bibliographystyle{elsarticle-harv}

\bibliography{refs}

\end{document}